\documentclass[10pt,twocolumn,letterpaper]{article}
\usepackage[utf8]{inputenc}
\usepackage[T1]{fontenc}
\usepackage{graphicx}
\usepackage{longtable}
\usepackage{wrapfig}
\usepackage{rotating}
\usepackage[normalem]{ulem}
\usepackage{amsmath}
\usepackage{amssymb}
\usepackage{capt-of}
\usepackage{hyperref}
\usepackage{authblk}
\usepackage{caption}
\usepackage[left=2.05cm,right=2.75cm,top=2.25cm,bottom=2.25cm]{geometry}
\usepackage{booktabs}
\usepackage{tabularx}
\usepackage{multirow}
\usepackage{dblfloatfix}
\usepackage{pdflscape}
\usepackage[utf8]{inputenc}
\usepackage{placeins}
\usepackage{etoc}
\author[1,2,3]{Jean-Baptiste Escudié}
\author[2]{Benjamin Barnes}
\author[2]{Stefan Meisegeier}
\author[2]{Klaus Kraywinkel}
\author[3]{Fabian Prasser}
\author[1]{Nils Körber\thanks{Corresponding author: koebern@rki.de.}}
\affil[1]{Centre for Artificial Intelligence in Public Health Research, Robert Koch Institute}
\affil[2]{German Centre for Cancer Registry Data, Robert Koch Institute}
\affil[3]{Medical Informatics, Berlin Institute of Health at Charité - Universitätsmedizin Berlin}
\date{}
\title{Evaluating quality in synthetic data generation for large tabular health datasets}
\hypersetup{
 pdfauthor={},
 pdftitle={Evaluating quality in synthetic data generation for large tabular health datasets},
 pdfkeywords={},
 pdfsubject={},
 pdfcreator={Emacs 30.2 (Org mode 9.7.11)}, 
 pdflang={English}}
\usepackage[style=apa]{biblatex}
\begin{document}

\maketitle
\begin{abstract}

There is no consensus in the field of synthetic data on concise metrics for quality evaluations or benchmarks on large health datasets, such as historical epidemiological data.
This study presents an evaluation of seven recent models from major machine learning families. The models were evaluated using four different datasets, each with a distinct scale. To ensure a fair comparison, we systematically tuned the hyperparameters of each model for each dataset.
We propose a methodology for evaluating the fidelity of synthesized joint distributions, aligning metrics with visualization on a single plot.
This method is applicable to any dataset and is complemented by a domain-specific analysis of the German Cancer Registries' epidemiological dataset. The analysis reveals the challenges models face in strictly adhering to the medical domain.
We hope this approach will serve as a foundational framework for guiding the selection of synthesizers and remain accessible to all stakeholders involved in releasing synthetic datasets.

\end{abstract}
\section{Introduction}
\label{sec:org779b7b9}
Synthetic tabular data is seeing a growing interest in conjunction to the increasing need for privacy preserving access to sensitive health data.
The two primary applications for synthetic health data are privacy and data augmentation. In the context of privacy preserving synthetic data, the promise is to be able to perform statistical analysis with similar outcome on an adjacent dataset that does not allow - or with a limited probability - to reindentify individuals, reconstruct or infer individual attributes from the original dataset. The quality of the synthesized datasets is a critical factor in both applications, as it is essential to ensure the fidelity of the datasets to the original data.

From classical statistical methods (copulas, Bayesian networks, SMOTE) tabular data synthesis evolved to deep generative models with the adaptation of VAEs and GANs for table (medGAN, table-GAN, CTGAN), capable of handling mixed-type columns and complex joint distributions, embeddings for categoricals and conditioning to stabilize discrete/continuous generation \autocite{choi_generating_2017,park_data_2018,xu_modeling_2019-1}. These deep models have then seen refined architectures and objectives proposed (CTAB-GAN, VAE variants, \autocite{zhao_ctab-gan_2021,zhao_ctab-gan_2023}), as well as differential privacy (DP) and Private Aggregation of Teachers Ensemble (PATE) based privacy variants introduced \autocite{yoon_pate-gan_2019}. More recent diffusion/score-based models and transformer/LLM-style sequence encodings have also been adapted for tabular synthesis \autocite{kotelnikov_tabddpm_2023,borisov_language_2023,truda_generating_2023}.

\begin{figure*}[!t]
  \centering
  \includegraphics[width=\linewidth]{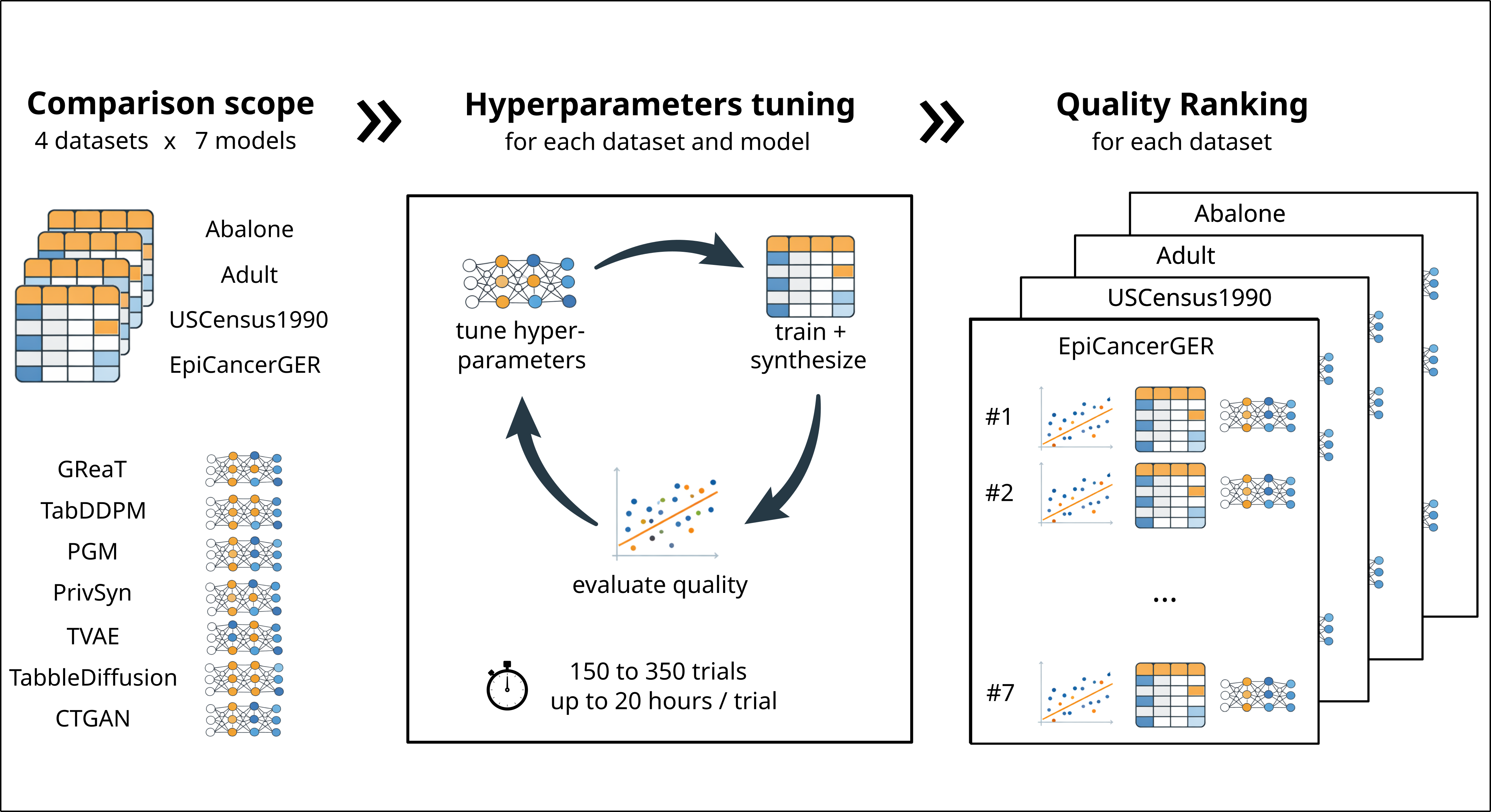}
  \caption*{Figure 1: Quality evaluation methodology overview.}
  \label{fig:Figure Overview}
\end{figure*}

Given the growing number of methods for synthesizing data that are being published, a comparison of these methods is a challenging task. The evaluation of synthetic data is a multifaceted process and the absence of consensus regarding the most suitable evaluation methods is noteworthy \autocite{lautrup_syntheval_2025}. The prevalence of metrics with minimal overlap in publications illustrates this phenomenon.

A categorization of the facets commonly evaluated is possible. It is common practice to differentiate between metrics that assess the quality (also called fidelity or resemblance) of the synthetic data, its utility, and its privacy. However, within each of these branches, a multitude of metrics exists. Taxonomies have been proposed \autocites[,]{kaabachi_scoping_2025}{hernandez_comprehensive_2025}{drechsler_30_2023}{vallevik_can_2024} and if they differ in structure and naming, they are designed to encompass the diverse metrics found in the literature in these three accepted categories of quality, utility and privacy. In many cases, a custom compound metric is constructed - e.g. by means of a weighted sum of seleted metrics  \autocite{chundawat_universal_2024,adams_fidelity_2025,yan_multifaceted_2022,du_systematic_2025-1} - to generate a single score.
Also, software libraries, e.g. \autocite{noauthor_synthetic_2025,qian_synthcity_2023}, collect large numbers of metrics implementations in one place. While this provides ready-to-use implementations, the interpretation of the results remains complex and context dependent.
In the context of searching for a privacy-preserving synthesizer suitable for a specific use case, it is imperative to evaluate the trade-off between maximizing privacy guarantees and ensuring the quality and utility of the synthesizer. The acceptability of particular choices is contingent upon the particulars of the context. In an ideal scenario, this process would entail the involvement of all relevant stakeholders, including patients, healthcare providers, public health agencies, epidemiologists, and researchers, irrespective of their statistical literacy. Consequently, there is a necessity for a straightforward method to assess quality, thereby providing guidance in determining the privacy guarantee tradeoff.

\begin{table*}[!t]
  \centering
  \begin{tabular}{|l|r|r|r|r|r|r|}
\toprule
Dataset & \multicolumn{2}{c|}{Categorical} & \multicolumn{1}{c|}{Numerical} & \multicolumn{3}{c|}{Total} \\
Name & N Variables & Categories & N Variables & N Variables & N Samples & Vector Size \\
\midrule
Abalone & 1 & 3 & 8 & 9 & 4175 & 11 \\
Adult & 9 & 106 & 6 & 15 & 48800 & 112 \\
EpiCancerGER & 10 & 718 & 0 & 10 & 1000000 & 718 \\
USCensus1990 & 68 & 396 & 0 & 68 & 2458200 & 396 \\
\bottomrule
\end{tabular}

  \caption*{Table 1: Datasets characteristics. Vector size denotes the size of a vector for a sample in the dataset after one-hot encoding the categorical variables.}
  \label{tab:Table 1}
\end{table*}

\begin{table*}[!b]
  \centering
  \begin{tabular}{|l|l|l|l|}
\toprule
Model & Family & Differential Privacy ability & Reference \\
\midrule
TVAE & Variational Auto Encoder & No & \cite{xu_modeling_2019-1} \\
CTGAN & GAN & No & \cite{xu_modeling_2019-1} \\
PGM & Probabilistic graph model & Yes & \cite{mckenna_graphical-model_2019} \\
PrivSyn & Distribution free model & Yes & \cite{zhang_privsyn_2021} \\
TabDDPM & Diffusion model & No & \cite{kotelnikov_tabddpm_2023} \\
TableDiffusion & Diffusion model & Yes & \cite{truda_generating_2023} \\
GReaT & Fined-tuned LLM & No & \cite{borisov_language_2023} \\
\bottomrule
\end{tabular}

  \caption*{Table 2: Models included in the benchmark.}
  \label{tab:Table 2}
\end{table*}

A further limitation in the comparison of published models stems from the heterogeneity of the datasets utilized, which frequently exhibit limited size, rarely exceeding 100,000 samples and a modest number of features, as seen in public machine learning datasets.
This prompts the inquiry into whether a reported method exhibits equivalent efficacy on more realistic datasets comprising a greater number of samples and/or features. Indeed, as the dataset under consideration grows, the resource requirements to apply a given method (runtime, memory) may become unrealistic. Furthermore, the hyperparameters that demonstrate efficacy on a specific dataset may exhibit suboptimal performance on a substantially distinct dataset.

We examine the synthesis of epidemiological data from the German Cancer Registries, which commenced in the 1970s and currently encompasses over 13 million cases. The data encompasses variables that encode clinical diagnoses in the International Classification of Diseases 10th Revision German Modification (ICD-10-GM). The dimensionality of medical classifications, such as ICD, often poses a significant challenge for machine learning models.

The present work focuses on the evaluation of quality from small to large tabular datasets. This benchmark is a comparative analysis of seven models representing recent major machine learning (ML) families. The models are evaluated on four different datasets, each with increasing dimensions in terms of number of rows and columns. The analysis has a particular focus on categorical data. The objective of this study is to establish a definitive foundation for the discussion of quality when selecting a method for synthesizing a privacy-preserving synthetic dataset.
To this end, hyperparameter optimization (HPO) was performed on each model and dataset to ensure a fair comparison. Subsequently, a simplified visual evaluation is presented, employing a limited yet sufficient number of metrics to rank the models and assess their performance.

The main contributions of this study are:
\begin{itemize}
\item a. We conduct a comprehensive scaling analysis, evaluating model performance across datasets of up to several million entries, which exceeds the scales typically reported.
\item b. Addressing the common issue of relying on default settings, this work provides an extensive empirical study on the impact of systematic tuning on the ability of models to converge to optimal hyperparameters.
\item c. Evaluation and visualisation methods that are concise yet enough to rank models, and can be extended for domain specific analysis.
\end{itemize}
\section{Methods}
\label{sec:org5fc13e2}

After an initial overview of the hyperparameter optimization and ranking, we present in the following sections the datasets, the models, the particularities of the HPO settings, the visualisations and metrics used for evaluation and ranking the models. We then conclude with the domain specific analysis extending the evaluation for the German Cancer Registries dataset.
\subsection{Overview}
\label{sec:org9f78681}

Given a dataset \(D_{real}\), a model \(M\) is trained and then used to generate a synthetic dataset \(D_{synth_{i}}\). The quality of \(D_{synth_{i}}\) compared to \(D_{real}\) is then measured.
Subsequently, another set of the model's hyperparameters is then tested and the process is reiterated until the tuning budget is depleted.

Upon completion of the independent tuning process for each combination of \(D_{real}\) and \(M\), the dataset \(D_{synth_{best}}\) is considered for the final ranking of models. A separate ranking of the models was performed for each dataset \(D_{real}\).
\subsection{Datasets}
\label{sec:org9d4caff}

The benchmark includes four datasets, as detailed in Table 1. Abalone \autocite{nash_abalone_1994} and Adult \autocite{becker_adult_1996} are two public datasets frequently utilized in machine learning with modest dimensions and heterogeneous types of variables (categorical and numerical).
EpiCancerGER is a subset of historical epidemiological data from all the cancer registries in Germany \autocite{robert_koch_institute_cancer_2023}, spanning from 1970 to 2019. This subset includes a total of one million cases. The list of included variables can be found in the Appendix.
The USCensus1990 \autocite{meek_us_2001} is a large-scale public dataset that contains exclusively categorical data.

\begin{table*}[!tb]
  \centering
  \begin{tabular}{|l|r|r|r|r|r|}
\toprule
{Dataset} & \multicolumn{2}{c|}{HPO budget} & \multicolumn{2}{c|}{HPO metrics} \\
{} & {N completed trials} & {Timeout} & {Categorical} & {Numerical} \\
\midrule
USCensus1990 & 150 & 10h + 10h & MAE2 &  \\
EpiCancerGER & 250 & 5h +  5h & MAE2 &  \\
Adult & 350 & 5h +  5h & MAE2 & Hist\_IoU2 \\
Abalone & 350 & 5h +  5h &  & Hist\_IoU2 \\
\bottomrule
\end{tabular}

  \caption*{Table 3: HPO settings. Timeout in HPO budget is split into training and synthesizing steps.}
  \label{tab:Table 3}
\end{table*}

This selection exhibits the range from 4k to 2.5M samples, and the encoded sample dimensions range from 11 to 540. It shoud be noted that missing values were considered to be an inherent category to be reproduced in the synthetic dataset. For this reason, the missing data is not reported seperately.
\subsection{Models}
\label{sec:orga1b7f03}

7 models were included in the benchmark as presented in Table 1. GReaT \autocite{borisov_language_2023} was included for reference, however due to its substantial computation costs, no HPO was conducted on it, and only a limited subset of 10k samples was utilized and generated.

A number of models - PGM \autocite{mckenna_graphical-model_2019}, PrivSyn \autocite{zhang_privsyn_2021}, TableDiffusion \autocite{truda_generating_2023} - have differential privacy capabilities integrated into their design. For these models, the epsilon value was set to 1e+8 to simulate a privacy budget that is nearly infinite. This was done to ensure a fair comparison against all models.

Implementations were retrieved from SynMeter \autocite{du_systematic_2025-1}, which aggregates either the original repositories or the SynthCity \autocite{qian_synthcity_2023} versions and provides default hyperparameters after SynMeter performed HPO in a different setting.
\subsection{Hyperparameters Optimization}
\label{sec:orgf5f3c94}

Table 3 details the characteristics of the HPO process that differed for each of the 4 datasets.

The ranges of hyperparameters are presented for each model in the [Appendix].

The rest of the HPO process remained the same for each combination of model and dataset. Namely, the Tree Parzen Estimator (TPE) \autocite{bergstra_algorithms_2011} was employed as optimizer for categorical only datasets (EpiCancerGER and USCensus1990). The HPO optimizer was the Multi Objective TPE \autocite{ozaki_multiobjective_2020} for mixed datasets (Abalone, Adult). The implementations of this methods were based on the Optuna library \autocite{akiba_optuna_2019}.

A salient feature of our HPO configuration was the stipulation that the number of trials should equal the number of completed trials. Consequently, in the event of a failed training or generation attempt, the model's budget did not account for the trial, and the HPO optimizer remained unaware of the failure. The motivation for this approach stems from the recognition that the causes of failures are heterogeneous. For instance, an invalid set of hyperparameters could have been sampled due to the presence of highly specific model requirements, such as the necessity of a batch size that is a multiple of two and of the PAC size. Furthermore, the implementation of a set of hyperparameters has the potential to result in a training or generation process that exhausts available resources, leading to errors such as timeout or out-of-memory exceptions. For the purpose of this benchmark, a selection of NVIDIA GPUs was utilized, including H100s, A100s, and L40s. Ultimately, it was not possible to attribute the failure entirely to the model implementations without compromising the fairness of the benchmark. To address this, the opportunity to complete the same number of trials with an unlimited number of failed trials was extended to every model.
\subsection{Evaluation metrics and visualisation}
\label{sec:org6ffe2e9}
For categorical variables, we primarily employed marginal distributional similarity, encompassing one-dimensional, pair-wise and n-way marginals. For numerical variables, for the main metric was the histogram's intersection over union for single and pairs of variables.
\subsubsection{Metrics for categorical variables}
\label{sec:org59af250}
The primary metric for categorical variables was derived from the pairwise marginals. This metric was employed in two distinct ways: first, to calibrate the HPO tuning, and second, to determine the final ranking of the models. In this study, the mean absolute error of all pairwise marginals, denoted \(MAE_2\), was utilized as the primary metric for analysis.

\begin{equation}\label{eq_MAE_2}
MAE_{2} = \frac{1}{|{V_{CAT}^{2}}|} * \sum_{v_{i},v_{j}}^{V_{CAT}^{2}} | p_{real}({v_{i},v_{j}}) - p_{synth}({v_{i},v_{j}}) |
\end{equation}

where \(V_{CAT}^{2}\) is the set of unordered pairs of categorical variables, \(p_{real}\) and \(p_{synth}\) are their empirical probabilities in the real and synthetic datasets respectively.

As secondary metrics, we also measured the single marginals and their corresponding mean absolute error \(MAE_{1}\) were measured. Two new metrics were defined: \(Coverage\) of categories for each variable and variables pairs, and the proportion of \(Invented\) relationships.

\(Coverage\) is defined per variable as the proportion of categories in \(D_{real}\) that are also generated at least once in \(D_{synth_{i}}\). To illustrate, if \(D_{synth_{i}}\) comprises solely samples with \(SEX=female\), yet \(D_{real}\) encompasses both \(SEX=male\) and \(SEX=female\) categories, then \(Coverage1_{SEX}\) value is determined to be 50$\backslash$%. The average of \(Coverage\) for every variable defines \(Coverage_{1}\), the coverage for all the variables the dataset.
Consequnetly, we defined \(Coverage_{2}\) as the coverage applied to every pair of variables.

\(Invented\) is the proportion of generated samples for which the value is not present in \(D_{real}\). Once more the average \(Invented_{1}\) for the dataset is the average for every variables.
\(Invented\) was measured only pairwise (\(Invented_{2}\)) as every model tested generates only valid categories (either because of the use of one-hot encoding or the application of filtering mechanisms to discard invalid values).
\subsubsection{Visualisations for categorical variables}
\label{sec:orgde7f8af}
Conveniently all these metrics can be repsented visually on two scatter plots: one for marginals, the second for the pair-wise marginals. \(MAE_{1}\) and \(MAE_{2}\) are the mean distance of every point from the diagonal line. The points falling on the \(x\) and \(y\) axis contributes to \(Coverage\) and \(Invented\) respectively.

\begin{figure*}[tb]
  \centering
  \includegraphics[width=\linewidth]{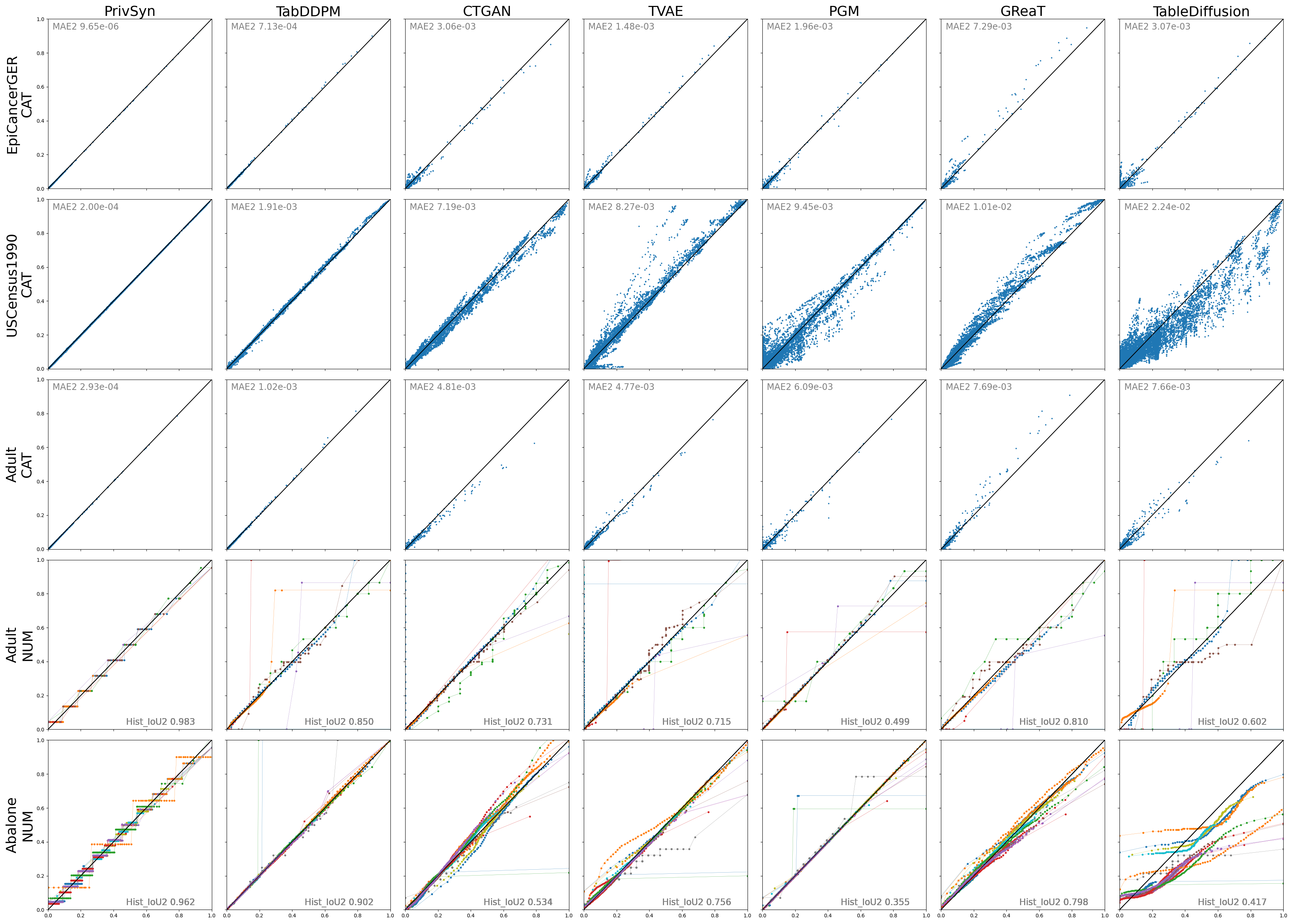}
  \caption*{Figure 2: Categorical (denoted CAT): scatter plots of pair-wise marginals for the best models after HPO. On each plot, a point is drawn for a pair-wise level. E.g. the point for SEX=MALE and ICD10=C20, is drawn at x=0.47 and y=0.55, if 47\% and 55\% are the observed percentages in the real and synthetic datasets respectively. Numerical (denoted NUM): QQ plots of every variable with the real quantiles on x-axis and the synthetic quantiles on the y-axis.}
  \label{fig:Figure 2}
\end{figure*}

\subsubsection{Metrics for numerical variables}
\label{sec:orga0decdc}
The primary metric for numerical variables was a binning distance on pairs of variables. This metric was employed both to calibrate the hyperparameters tuning and to determine the final ranking of the models. Specifically we discretized the numerical variable using 10 bins of equal sizes and computed the average intersection over union (IOU) of the corresponding histograms.

A similar metric was measured for variables instead of pairs. We'll refer to these metrics as \(Hist\_IoU_{1}\) and \(Hist\_IoU_{2}\). For visual interpretation, they conveniently correspond to the average L1 distance in all the 1D and 2D histograms constructed from the sets of variables and pairs of variables respectively. Theese values correspond as well to the mean absolute error of empirical probabilities following formula \eqref{eq_MAE_2} with \(V_{NUM_{bin}}^{2}\) as the set of unordered pairs of binned numerical variables.
\subsubsection{Visualisation for numerical variables}
\label{sec:org01c86a3}
Visualizing a set of 1D or 2D histograms is a non trivial task. Instead, quantile-quantile (QQ) plots are presented for all variables in a single plot, with each colored line representing a distinct variable.
\subsection{Domain violation}
\label{sec:org0cd7a39}
To complement the evaluation of the EpiCancerGER dataset, an assessment of domain violation in the synthetic dataset was conducted. This analysis is distinct from the other metrics in that it is dataset-specific. In order to execute this analysis, it is necessary to manually define the domain.

This analsysis sought to define domain violation for three pairs of variables involving ICD codes. In EpiCancerGER, both full and 3-digit ICD codes are included.

This enables the establishment of a primary domain that is unambiguously delineated. A violation is identified when the two codes do not correspond to the ICD classification system. Specifically, this occurs when the ICD's three-digit code does not serve as a prefix to the entire code. The synthesizer should be capable of recovering this strict matching, as it is strictly adhered to in the real dataset. However, given that both codes are represented as independent variables, it is necessary to assess whether the models respect the ICD structure strictly.

The second case of domain validation analysis was defined for the pair of variables sex and ICD 3-digits. Again, there exists a strict exclusion verified in the real dataset for 15 ICD chapters covering genital organs neoplasms listed in the [Appendix].

The third and final case of domain validation involved a comparison of age groups with ICD 3-digits codes. For this analysis, the minimum and maximum ages observed in the real dataset for each ICD chapter were determined. Subsequently, it was verified whether the synthetic dataset respects these established ranges. This is often referred as out-of-range detection.
\section{Results}
\label{sec:orgeabccd7}
\subsection{Main results - best models after HPO}
\label{sec:org9c34df4}

\begin{table*}[bt]
  \centering
  \resizebox{\textwidth}{!}{
    \begin{tabular}{|l|l|r|r|r|r|r|r|r|}
\toprule
{} & {} & \multicolumn{5}{c|}{\bfseries Categorical} & \multicolumn{2}{c|}{\bfseries Numerical} \\
{} & {} & {\bfseries MAE1 ↓} & {\bfseries MAE2 ↓} & {\bfseries Coverage1 ↑} & {\bfseries Coverage2 ↑} & {\bfseries Invented2 ↓} & {\bfseries Hist\_IoU1 ↑} & {\bfseries Hist\_IoU2 ↑} \\
{dataset} & {model} & {} & {} & {} & {} & {} & {} & {} \\
\midrule
\multirow[t]{7}{*}{USCensus1990} & PrivSyn & 0.0002 & \bfseries 0.0002 & \bfseries 1.0000 & 0.9828 & \bfseries 0.0000 &  &  \\
 & TabDDPM & 0.0034 & 0.0019 & \bfseries 1.0000 & 0.9916 & 0.0002 &  &  \\
 & CTGAN & 0.0121 & 0.0065 & \bfseries 1.0000 & \bfseries 1.0000 & 0.0053 &  &  \\
 & TVAE & 0.0129 & 0.0070 & 0.9931 & 0.9275 & 0.0005 &  &  \\
 & PGM & \bfseries 0.0000 & 0.0095 & \bfseries 1.0000 & 0.9981 & 0.0265 &  &  \\
 & GReaT & 0.0174 & 0.0101 & 0.9517 & 0.8184 & 0.0000 &  &  \\
 & TableDiffusion & 0.0385 & 0.0224 & \bfseries 1.0000 & \bfseries 1.0000 & 0.0225 &  &  \\
\midrule
\multirow[t]{7}{*}{EpiCancerGER} & PrivSyn & 0.0000 & \bfseries 0.0000 & 0.9795 & 0.9398 & 0.0001 &  &  \\
 & TabDDPM & 0.0019 & 0.0006 & 0.9975 & 0.9373 & 0.0003 &  &  \\
 & TVAE & 0.0041 & 0.0015 & 0.9881 & 0.9234 & 0.0068 &  &  \\
 & PGM & \bfseries 0.0000 & 0.0020 & \bfseries 1.0000 & 0.9670 & 0.0401 &  &  \\
 & TableDiffusion & 0.0073 & 0.0028 & 0.6950 & 0.4545 & 0.0062 &  &  \\
 & CTGAN & 0.0087 & 0.0031 & \bfseries 1.0000 & \bfseries 0.9813 & 0.0163 &  &  \\
 & GReaT & 0.0170 & 0.0073 & 0.8904 & 0.7339 & \bfseries 0.0000 &  &  \\
\midrule
\multirow[t]{8}{*}{Adult} & PrivSyn & 0.0005 & \bfseries 0.0003 & \bfseries 1.0000 & 0.9416 & 0.0002 & 0.0337 & \bfseries 0.9830 \\
 & TabDDPM & 0.0030 & 0.0009 & 0.9974 & 0.9283 & 0.0001 & \bfseries 0.9010 & 0.8931 \\
 & TVAE & 0.0125 & 0.0043 & 0.9974 & 0.9559 & 0.0011 & 0.6580 & 0.6976 \\
 & CTGAN & 0.0137 & 0.0043 & \bfseries 1.0000 & \bfseries 0.9889 & 0.0035 & 0.6148 & 0.7286 \\
 & PATE-GAN & 0.0148 & 0.0052 & 0.9259 & 0.8293 & 0.0092 & 0.6131 & 0.7482 \\
 & PGM & \bfseries 0.0000 & 0.0061 & \bfseries 1.0000 & 0.9758 & 0.0116 & 0.4004 & 0.4990 \\
 & TableDiffusion & 0.0283 & 0.0077 & 0.9071 & 0.8085 & 0.0003 & 0.6903 & 0.6019 \\
 & GReaT & 0.0239 & 0.0077 & 0.9277 & 0.7961 & \bfseries 0.0001 & 0.8438 & 0.8099 \\
\midrule
\multirow[t]{7}{*}{Abalone} & PrivSyn & 0.0010 &  & \bfseries 1.0000 &  &  & 0.0659 & \bfseries 0.9622 \\
 & TabDDPM & 0.0110 &  & \bfseries 1.0000 &  &  & \bfseries 0.8927 & 0.9142 \\
 & GReaT & 0.0096 &  & \bfseries 1.0000 &  &  & 0.8004 & 0.7984 \\
 & TVAE & 0.0762 &  & \bfseries 1.0000 &  &  & 0.7478 & 0.7711 \\
 & CTGAN & 0.0089 &  & \bfseries 1.0000 &  &  & 0.7454 & 0.5691 \\
 & TableDiffusion & 0.1851 &  & \bfseries 1.0000 &  &  & 0.4636 & 0.4440 \\
 & PGM & \bfseries 0.0000 &  & \bfseries 1.0000 &  &  & 0.6941 & 0.3547 \\
\bottomrule
\end{tabular}

  }
  \caption*{Table 4: Main results, best models after HPO. ↓ denotes lower is better. ↑ denotes higher is better. The $MAE_{2}$ and $Hist\_IoU_{2}$ metrics rank the models. Bold metrics values denotes the best in dataset.}
  \label{tab:Table 4}
\end{table*}

The comparison of the top-performing models following HPO reveals that PrivSyn emerged as the leading option across all datasets in terms of pairwise categorical metrics. This phenomenon is illustrated in the accompanying plots, where the points align more closely with the diagonal line, indicating that the real marginals equal the synthetic ones. This objective was accomplished with minimal disruption to existing pairwise relationships, while ensuring a large coverage of all existing relationships. PrivSyn also demonstrated optimal performance in the joint distribution of numerical variables in the Adult dataset. The QQ plot manifests a stepping effect, attributable to the optimal number of bins as determined by HPO, which in turn drives the binning performed internally by PrivSyn. The introduction of the artifact by the model is reflected in the conspicuously substandard \(Hist\_IoU_{1}\) metric.

The second-best model demonstrated a consistent pattern of performance across all datasets, with Tabbdpm exhibiting the second highest level of performance. The presence of a systematic artifact, such as PrivSyn, on the numerical variables was not observed. Consequently, the optimal alignment of 1D histograms with real data was achieved. A comparison of the metrics \(Invented_{2}\) and \(Coverage\) with those of PrivSyn revealed no significant disparities. However, a notable divergence was observed in the primary joint distribution metrics, specifically \(MAE_{2}\) and \(Hist\_IoU_{2}\).

The following models in the ranking were TVAE and CTGAN. It is noteworthy that the performance of the two models was comparable, with the exception of EpiCancerGER. On the two primary joint distribution metrics, \(MAE_2\) and \(Hist\_IoU_2\), the models exhibited divergent behaviors. In fact, the TVAE model exhibited a tendency to overestimate relationships as the prevalence increased. Conversely, the CTGAN model demonstrated a propensity to underestimate relationships, a phenomenon that is evident in the scatter plots. Additionally, TVAE demonstrated a preference for low \(Invented_{2}\) over high \(Coverage_{2}\), while CTGAN exhibited the inverse preference. Additionally, TVAE outliers, defined as variables synthesized with reduced quality, exhibit a greater disparity from the original data distribution. In contrast, CTGAN demonstrates a more uniform performance across variables. A more thorough examination determined that TVAE exhibited unambiguous artifacts, particularly evident in the USCensus1990 data. These artifacts manifested as lines that deviated from the primary body of points, suggesting a propensity to either over- or underestimate specific values in proportion to their support in the original data.

PGM and TableDiffusion demonstrated suboptimal performance in this benchmark, with PGM exhibiting superior performance compared to TableDiffusion. PGM demonstrated a notable capacity to accurately estimate the marginal probabilities; however, its performance was suboptimal in terms of capturing the joint distributions, particularly in scenarios where prevalence levels were lower.

\begin{figure*}[htbp]
  \centering
  \includegraphics[width=0.92777\linewidth, height=0.92777\textheight, keepaspectratio]{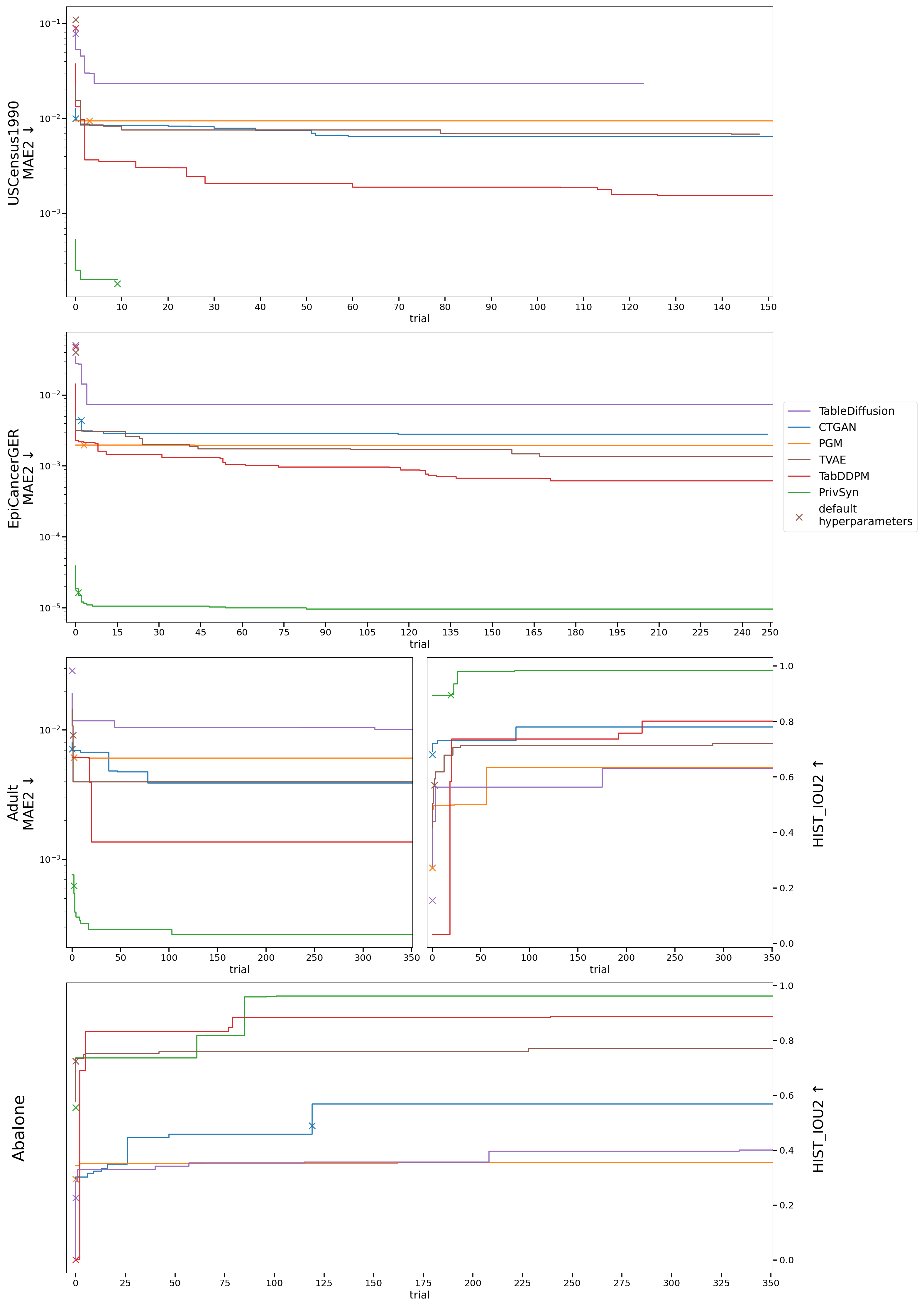}
  \caption*{Figure 3: Evaluation metrics over HPO tuning. The y-axis on the left represents $MAE_2$ and $Hist\_IoU_2$ on the right. Cross marks locate when the default hyperparameters were outperformed.}
  \label{fig:Figure 3}
\end{figure*}

\subsection{HPO improvement compared to default hyperparameters}
\label{sec:orge2819cd}

Table 5 compares the main metrics \(MAE_{2}\) and \(Hist\_IoU_{2}\) with the default hyperparameters provided in SynMeter \autocite{du_systematic_2025-1} (with the exception of relaxed differential privacy) and after the HPO tuning performed in this study. In the vast majority of cases, the HPO led to a significant improvement in quality for both numerical and categorical variables. On the two large datasets USCensus1990 and EpiCancerGER, for every models except PrivSyn, the default hyperparameters result in suboptimal performance. The most significant case is TabDDPM which demonstrated the most substantial enhancement of 85,797\%. In the absence of HPO, the model would have been considered substandard. However, it demonstrated a second-best performance in our primary ranking, indicating its reliance on hyperparameters choice.

\begin{figure*}[hbtp]
  \centering
  \includegraphics[width=0.9777\linewidth, height=0.9777\textheight, keepaspectratio]{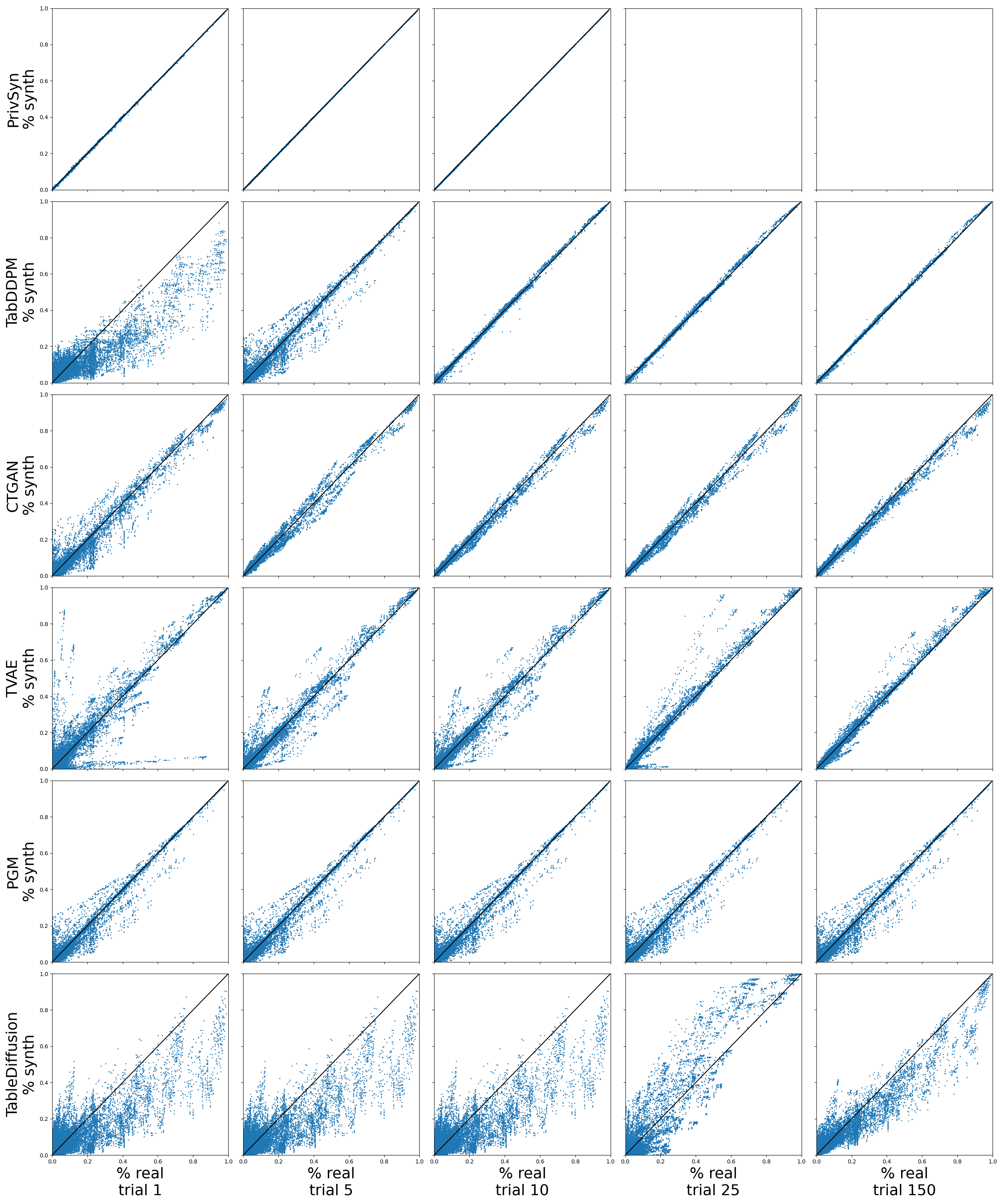}
  \caption*{Figure 4: Evaluation pair-wise scatter plots over the HPO tuning for USCensus1990.}
  \label{fig:Figure 4}
\end{figure*}

As demonstrated in Figure 3, the maximum perfomance of most models was typically attained during the first 150 HPO trials.

\begin{table*}[t]
  \centering
  \resizebox{0.777\textwidth}{!}{
    \begin{tabular}{|l|l|r|r|r|r|r|r|r|r|}
\toprule
{} & {} & \multicolumn{4}{c|}{\bfseries Categorical MAE2 ↓ } & \multicolumn{4}{c|}{\bfseries Numerical Hist\_IoU2 ↑ } \\
{} & {} & {\bfseries default} & {\bfseries HPO} & \multicolumn{2}{c|}{\bfseries improvement} & {\bfseries default} & {\bfseries HPO} & \multicolumn{2}{c|}{\bfseries improvement} \\
{} & {} & {\bfseries } & {\bfseries } & {\bfseries } & {\bfseries \%} & {\bfseries } & {\bfseries } & {\bfseries } & {\bfseries \%} \\
{dataset} & {model} & {} & {} & {} & {} & {} & {} & {} & {} \\
\midrule
\multirow[t]{7}{*}{USCensus1990} & PrivSyn & 0.0002 & 0.0002 & 0.0000 & 10\% &  &  &  &  \\
 & TabDDPM & 0.0891 & 0.0019 & -0.0872 & -98\% &  &  &  &  \\
 & CTGAN & 0.0100 & 0.0065 & -0.0035 & -35\% &  &  &  &  \\
 & TVAE & 0.1094 & 0.0070 & -0.1024 & -94\% &  &  &  &  \\
 & PGM & 0.0095 & 0.0095 & -0.0000 & -0\% &  &  &  &  \\
 & GReaT & 0.0101 & N/A & N/A & N/A &  &  &  &  \\
 & TableDiffusion & 0.0780 & 0.0224 & -0.0556 & -71\% &  &  &  &  \\
\midrule
\multirow[t]{7}{*}{EpiCancerGER} & PrivSyn & 0.0000 & 0.0000 & -0.0000 & -41\% &  &  &  &  \\
 & TabDDPM & 0.0475 & 0.0006 & -0.0469 & -99\% &  &  &  &  \\
 & TVAE & 0.0400 & 0.0015 & -0.0385 & -96\% &  &  &  &  \\
 & PGM & 0.0020 & 0.0020 & -0.0000 & -0\% &  &  &  &  \\
 & TableDiffusion & 0.0504 & 0.0028 & -0.0476 & -94\% &  &  &  &  \\
 & CTGAN & 0.0044 & 0.0031 & -0.0013 & -30\% &  &  &  &  \\
 & GReaT & 0.0073 & N/A & N/A & N/A &  &  &  &  \\
\midrule
\multirow[t]{8}{*}{Adult} & PrivSyn & 0.0006 & 0.0003 & -0.0003 & -53\% & 0.8955 & 0.9830 & 0.0876 & 10\% \\
 & TabDDPM & 0.0007 & 0.0009 & 0.0003 & 39\% & 0.8901 & 0.8931 & 0.0031 & 0\% \\
 & TVAE & 0.0091 & 0.0043 & -0.0048 & -53\% & 0.5705 & 0.6976 & 0.1271 & 22\% \\
 & CTGAN & 0.0071 & 0.0043 & -0.0028 & -39\% & 0.6806 & 0.7286 & 0.0480 & 7\% \\
 & PATE-GAN & 0.0254 & 0.0052 & -0.0202 & -79\% & 0.3405 & 0.7482 & 0.4076 & 120\% \\
 & PGM & 0.0061 & 0.0061 & -0.0000 & -1\% & 0.2723 & 0.4990 & 0.2267 & 83\% \\
 & TableDiffusion & 0.0288 & 0.0077 & -0.0211 & -73\% & 0.1551 & 0.6019 & 0.4468 & 288\% \\
 & GReaT & 0.0077 & N/A & N/A & N/A & 0.8099 & N/A & N/A & N/A \\
\midrule
\multirow[t]{7}{*}{Abalone} & PrivSyn &  &  &  &  & 0.5563 & 0.9622 & 0.4058 & 73\% \\
 & TabDDPM &  &  &  &  & 0.0011 & 0.9142 & 0.9131 & 85,797\% \\
 & GReaT &  &  &  &  & 0.7984 & N/A & N/A & N/A \\
 & TVAE &  &  &  &  & 0.7250 & 0.7711 & 0.0461 & 6\% \\
 & CTGAN &  &  &  &  & 0.4894 & 0.5691 & 0.0797 & 16\% \\
 & TableDiffusion &  &  &  &  & 0.2262 & 0.4440 & 0.2179 & 96\% \\
 & PGM &  &  &  &  & 0.2953 & 0.3547 & 0.0594 & 20\% \\
\bottomrule
\end{tabular}

  }
  \caption*{Table 5: HPO improvement. Comparing metrics after HPO to default hyperparameters.  ↓ denotes lower is better. ↑ denotes higher is better.}
  \label{tab:Table 5}
\end{table*}

PGM and PrivSyn exhibited minimal variations in performance across the various trials and demonstrated reduced reliance on their hyperparameters.
\subsection{Impact of dataset size}
\label{sec:orgdbdce74}
Due to the dataset-dependence of our metrics, we employed models ranking, and visual interpretation to estimate the impact of dataset sizes on model performance. Consequently, it is challenging to evaluate the impact of dataset size on behavior based on dataset size for PrivSyn and TabDDPM as these two models consistently achieved the top two rankings. However, a more substantial enhancement was observed for \(PrivSyn\) following HPO tuning for the smaller datasets, Adult and Abalone; this was in contrast to the observations for USCensus1990 and EpiCancerGER. The significance of HPO for TabDDPM appears to be independent of dataset size. However, an exception was observed for the Adult dataset, where TabDDPM demonstrated a relative decline in performance following HPO.
\subsection{Domain violation}
\label{sec:org672f36d}

\begin{table*}[hbpt]
  \centering
  \resizebox{\textwidth}{!}{
    \begin{tabular}{|l|r|r|r|r|r|r|}
\toprule
{} & \multicolumn{2}{c|}{\bfseries ICD vs ICD 3-digit} & \multicolumn{2}{c|}{\bfseries Sex vs ICD 3-digit} & \multicolumn{2}{c|}{\bfseries Age group vs ICD 3-digit} \\
{} & \multicolumn{2}{c|}{\bfseries 60,048 possible levels (556 x 108) } & \multicolumn{2}{c|}{\bfseries 216 possible levels (2 x 108)   } & \multicolumn{2}{c|}{\bfseries 1,944 possible levels (18 x 108)  } \\
{} & \multicolumn{2}{c|}{\bfseries 556 levels observed in real data} & \multicolumn{2}{c|}{\bfseries 201 levels observed in real data} & \multicolumn{2}{c|}{\bfseries 1,745 levels observed in real data} \\
{} & \multicolumn{2}{c|}{\bfseries } & \multicolumn{2}{c|}{\bfseries } & \multicolumn{2}{c|}{\bfseries } \\
{} & \multicolumn{2}{c|}{\bfseries Domain violations} & \multicolumn{2}{c|}{\bfseries Domain violations} & \multicolumn{2}{c|}{\bfseries Domain violations} \\
{} & {\bfseries n distinct} & {\bfseries \% samples} & {\bfseries n distinct} & {\bfseries \% samples} & {\bfseries n distinct} & {\bfseries \% samples} \\
\midrule
GReaT & 0 & 0.00\% & 0 & 0.00\% & 0 & 0.00\% \\
PrivSyn & 761 & 0.08\% & 8 & 0.00\% & 46 & 0.01\% \\
TabDDPM & 4,889 & 1.07\% & 15 & 0.02\% & 56 & 0.01\% \\
TVAE & 2,559 & 21.13\% & 11 & 0.19\% & 49 & 0.01\% \\
TableDiffusion & 598 & 24.93\% & 3 & 0.02\% & 0 & 0.00\% \\
CTGAN & 39,379 & 46.41\% & 15 & 0.57\% & 84 & 0.03\% \\
PGM & 27,953 & 93.48\% & 15 & 7.97\% & 92 & 0.06\% \\
\bottomrule
\end{tabular}

  }
  \caption*{Table 7: Domain violations on EpiCancerGER dataset. Number of distinct level pairs outside of domain and percentage of samples affected in the synthetic data.}
  \label{tab:Table 7}
\end{table*}

As demonstrated in Table 7, all models exhibited domain violations in the EpiCancerGER dataset, though to varying extents. It was evident that all models generated invalid pairs of values; however, the coverage of the invalid possible space varied significantly. For instance, the range of ICD vs. ICD 3-digits varied from 598 (1.01\%) of the 59,492 possible violations (60,048 possible pairs - 556 pairs valid and present in the real dataset) for TableDiffusion to 39,379 (60.2\%) for CTGAN. The proportion of the 1 million synthesized samples affected was not systematically proportional to this coverage of invalid code space: TabDDPM exhibited a mere 1.07\% of the samples to be in violation of the ICD code chapters structure, yet still encompassed 8.22\% of the potential violations.

Over the 3 ICD domain violations we assessed, PrivSyn exhibited the lowest number of affected samples, while PGM demonstrated the highest.
\section{Discusion}
\label{sec:org361641b}
We performed a quality evaluation of the datasets generated by 7 recently published models of different major ML families, after a systematic hyperparameters tuning to each of the four datasets of different scales. We have developed a novel evaluation method that utilizes a concise set of metrics, aligned with a condensed visual representation. This method was sufficiently comprehensive for the purpose of model ranking. We have demonstrated on the EpiCancerGER dataset how it can be extended by additional domain-specific analysis.
\subsection{Evaluation metrics choice}
\label{sec:org3362713}
In order to address the paucity of consensus on evaluation metrics in the discipline of synthetic data, while hopefully providing an approach that is sufficiently simple to be comprehensible to all stakeholders involved in the process of releasing a synthetic dataset, regardless of their statistical literacy, we employed marginal-based metrics \(MAE\), \(Coverage\), and \(Invented\).
Marginal statistics are frequently employed for quality assessment, yet they lack the capacity to measure joint distributions. Relationships constitute an indispensable component of the data to be reproduced in the synthesized datasets. Consequently, the metrics were defined on the independent variable (degree 1) and every pair of variables (degree 2).

The closest existing metric to \(MAE_{1}\) is the total variation distance (TVD), with a value that is half of \(MAE_{1}\). Being an average, it may be insufficient to analyze the case when differences with the original marginals are the most extreme. To this end, we defined \(Coverage\) and \(Invented\). We observed a tradeoff between \(Coverage\) on one end and \(MAE\) and \(Invented\) on the other end. The precise mechanism is contingent upon the internals of each model; however, a parallel can be drawn between this tradeoff and an exploration-exploitation tradeoff in the context of GAN models. A mode collapse occurs when the model exclusively generates the most prevalent categories, also known as the mode. This phenomenon constitutes a local minimum that is nonetheless an effective strategy, wherein the generator utilizes a salient aspect of the original data to confound the discriminator. This scenario would result in a reduction of the metrics related to coverage. Conversely, in the scenario of exhaustive exploration, the model would not acquire the capacity to restrict the extent of generated value to the existing pairs of categories, thereby augmenting the \(Invented2\) metric.

In this particular benchmark, the emphasis was placed on categorical data. However, it was demonstrated that \(MAE\) can be adapted to numerical variables through the process of binning, resulting in the \(Hist\_IoU\) metric.
\subsection{Visualization}
\label{sec:org84fa377}
Scatter plots can serve as a graphical method for comparing the distributions of categorical variables in a single plot. However, they are seldom employed aggregated at the level of the dataset for all variables, and this is typically only done for the purpose of visualizing independent distributions \autocite{baowaly_synthesizing_2019}. We introduced scatter plots on every pair of variables, which are utilized for the visual assessment of the joint distributions. The scatterplots demonstrate a strong correlation with the \(MAE\), \(Coverage\), and \(Invented\) metrics, each of which has a readily interpretable graphical representation. \(MAE\) is the average distance to the identity line which corresponds the ideal case where all synthetic marginals match the real ones and mimimize the metric to zero. \(Coverage\) is the proportion of points falling the x-axis. And \(Invented\) is the proportion of generated samples with the corresponding marginals falling on the y-axis.

In the case of numerical variables, a graphical interpretation of \(Hist\_IoU\) is also applicable. It is the average of the L1 distances of all 1D histograms (\(Hist\_IoU_{1}\)) or 2D histograms (\(Hist\_IoU_{2}\)).

The visualization in one plot of the comparison of all numerical independent or joint distributions in the dataset remains a challenging endeavor, necessitating further investigation and refinement of analytical methods. A common approach relies on plotting two heatmaps side by side of the correlation matrices, as illustrated in \autocite{rohrig_nfdi4health_2024}. It requires choosing appropriate correlation measures and does not scale for large number of variables. We instead opted to present QQ plots, which have the capacity to represent all independent variables (one line per variable) in a single plot. However, this approach is not without its limitations, as it does not extend to joint distributions.
\subsection{Relation to other metrics}
\label{sec:org243c9b9}
Although covering joint distributions, our marginal based metrics yielded rankings of the models that may differ if other metrics were to be used.

A comparison was made between the rankings obtained using the Jensen-Shannon distance (JSD) and the Wasserstein distance (WD), which are also commonly used in synthetic data evaluation. If the relationships between any pair of the \(MAE_{2}\), JSD or WD measurements is not strictly monotonic, the ranking would still have been unchanged for the first 4 best models and the least performing as illustrated in Figure 5 in the Appendix.
\subsection{Domain violations analysis}
\label{sec:org7c6f01b}
The generic quality evaluation methods proposed herein have the capacity to be complemented in order to address the specifics of the dataset. In the case of the EpiCancerGER dataset, the structure and semantics of the International Classification of Diseases (ICD) were leveraged to define domains with clearly delineated boundaries of invalid values. Domain driven definitions for exact boundaries can be challenging, and two alternative approaches were identified in our literature review. \autocite{yan_multifaceted_2022} leveraged a data-driven approach identifying diseases present only for one gender in the real data. Rather than being a external definition of the domain, it is in fact closer to the notion of fidelity to the original data and to the Invented metric, and does not guarantee domain correctness. A typical counter example would be breast and prostate cancers: men with breast cancer could be absent of the original data but still be valid if synthesized, whereas women with prostate cancer would be invalid. \autocite{pilgram_magnitude_2025} delineates clearly these two cases in their definition of hallucinations in synthetic tabular data and propose in their work another data-driven approach to define the source of gound truth from the data by creating variant of the data population and training the models on only a subset. A record is considered not hallucinated if present in the source population, whether present or not in the training data.

Our preliminary investigation indicated a challenge faced by all models except GReaT in strictly adhering to a medically coherent domain. To varying extents, each model yielded full ICD codes that did not align with the ICD chapters, or ICD chapters that were incompatible with the sex or age group attributed to the synthesized sample. In the case of GReaT, the model did not invent any relationship on the EpiCancerGER dataset - thus the absence of domain violations - but at the cost of significantly lower \(Coverage_2\) and \(MAE_{2}\). It is noteworthy that, in our experiments, domain violations increase with the cardinality of the variable pair. This aligns with \autocite{pilgram_magnitude_2025} findings, despite their different definition. They observed that the main factor driving hallucination rates was dataset cardinality.
Depending on the application, strategies to mitigate domain violations could include the filtration of invalid samples. However, this approach is not without its drawbacks, as it may result in interaction with the measured marginals and the potential introduction of underepresentation bias if the filtered out incoherences are not independent of other patterns in the data. Therefore, such mitigation requires careful consideration.
\subsection{Effect of HPO}
\label{sec:org44ecd73}
Given the models' high sensitivity to their hyperparameters, with the exception of PGM, it is recommended to tune their values for each dataset. This recommendation aligns with the approach outlined in \autocite{du_systematic_2025-1}, which advocates for the utilization of an HPO algorithm, albeit with three key diffrerences. Firstly, the objective of tuning differs because their emphasis is on the application preserving privacy synthetic data, rather than focusing quality. The researchers employed a multifaceted approach, leveraging a weighted sum of membership disclosure for privacy, query error for utility, and Wasserstein distance for fidelity score. The weighting of these three components was empirically set to equal weights after observing similar ranges for the three components. The optimization of such a compound objective may result in the prioritization of one objective over the other objectives. A similar challenging situation was encountered in our study with the Adult dataset, which has both numerical and categorical variables and necessitated the implementation of a Pareto front to ascertain the optimal trials. This approach is outlined in Optuna's multi-objective feature \autocite{akiba_optuna_2019}. For the other datasets, our objective was more straightforward: we sought to identify a single metric (either \(MAE_2\) or \(Hist\_IoU_2\)).

A second key aspect of the methodology that was implemented differently was the use of a time budget and allowance for unlimited trials, but only counting completed trials in the models HPO budget. As delineated in the Methods section, it is posited that this approach engenders a more equitable comparison. The rationale underlying this assertion is that subjecting the models to a broad spectrum of parameters and dataset scales, which may not have been encountered during their development, can precipitate either the manifestation of bugs or the exhaustion of resources (e.g., RAM, GPU RAM, runtime).
By not penalizing the implementations we tested for not being error proof on our datasets, or having trials failing because they ran on the most modest of our mix of GPUs, our unlimited trials HPO settings actively compensated for these phenomenons.

The third different key aspect was the number of trials conducted for the HPO. According to the published source code of SynMeter, 50 trials were utilized. We used 350 trials for the datasets we share with their experiments (Adult and Abalone), and 250 and 150 for our larger datasets (EpiCancerGER and USCensus1990, respectively). In our experiments this proved to be beneficial as metrics kept improving up to the 240th trial for TabDDPM. Although this step of last observed improvement varied with models and datasets. It is also noteworthy that the change in hyperparameters had a negligible impact on PGM performance. Finally, one may consider the cost versus marginal improvement of longer HPO.
\subsection{Bests models}
\label{sec:org43a1a54}
PrivSyn was found to consistently yield optimal quality metrics (\(MAE_{2}\), \(Hist\_IoU_{2}\)) across datasets in the context of the benchmark, while concomitantly maintaining adequate coverage of pairwise joint categorical distributions (\(Coverage_{2}\)).
Depending on the application (e.g., data augmentation, privacy), concerns about generating a synthetic dataset too close to the original may be raised. In such cases, a dedicated analysis would be required. PrivSyn was developped with builtin differential privacy for the privacy preserving synthetic data application. In the context of our experimental setup, the epsilon parameter was configured to 1e+8. This configuration was implemented with the objective of facilitating a meaningful comparison in terms of quality with other models while concurrently minimizing the impact of the differential privacy feature. How the quality degrades as the epsilon is lowered to acceptable values in privacy application could guide the decision process regarding the quality/privacy tradeoff.

PrivSyn's capacity to generate data close to the original without inventing relationships between variables is also an important aspect in the application of data augmentation. But our quality analysis should be complemented with other aspects such as measuring the capacity to generate close but non existing samples, quantified with the Distance to Closest Record (DCR) for example.

TabDDPM emerged as the second most effective model across all datasets. As with PrivSyn, the \(Coverage\) was satisfactory, though not the optimal values were observed. Furthermore, the invented relationships were found to be at a notably low level. This outcome is in stark contrast with the optimal \(Coverage_{2}\) values observed for CTGAN (on Adult and EpiCancerGER) and TableDiffusion (on USCensus1990). However, this is achieved at the expense of a substantial increase in \(Invented_{2}\) and a deterioration in \(MAE_{2}\).
\section{Conclusion}
\label{sec:orga4294e5}
In this study focusing on quality of synthetic data, we benchmarked 7 models on four datasets of varying scales in both number of samples and number of variables. In order to ensure a fair comparison, the challenge of models sentivity to hyperparameters was overcome by means of a systematic tuning of each model to each dataset. The best models after tuning were compared and ranked using a limited set of metrics. The proposed visualization method facilitates the interpretation of distances to pair-wise margins, coverage of joint distributions, and the propensity of a model to invent relationships absent in real data, all within a single plot. A demonstration was conducted on a German Cancer Registries dataset to illustrate how application-specific analysis can complement the proposed generic quality assessment of synthetic tabular data. We hope these methods provide a clear foundation for discussing quality when choosing a method for synthesizing a privacy preserving synthetic dataset.
\section{References}
\label{sec:org53c0b3d}
\printbibliography

\onecolumn
\clearpage
\section{Appendix}
\label{sec:orga86f425}
\appendix
\localtableofcontents

\clearpage
\subsection{Comparison of rankings from different metrics}
\label{sec:org7e702f2}

\begin{figure*}[h!]
  \centering
  \includegraphics[width=0.777\linewidth]{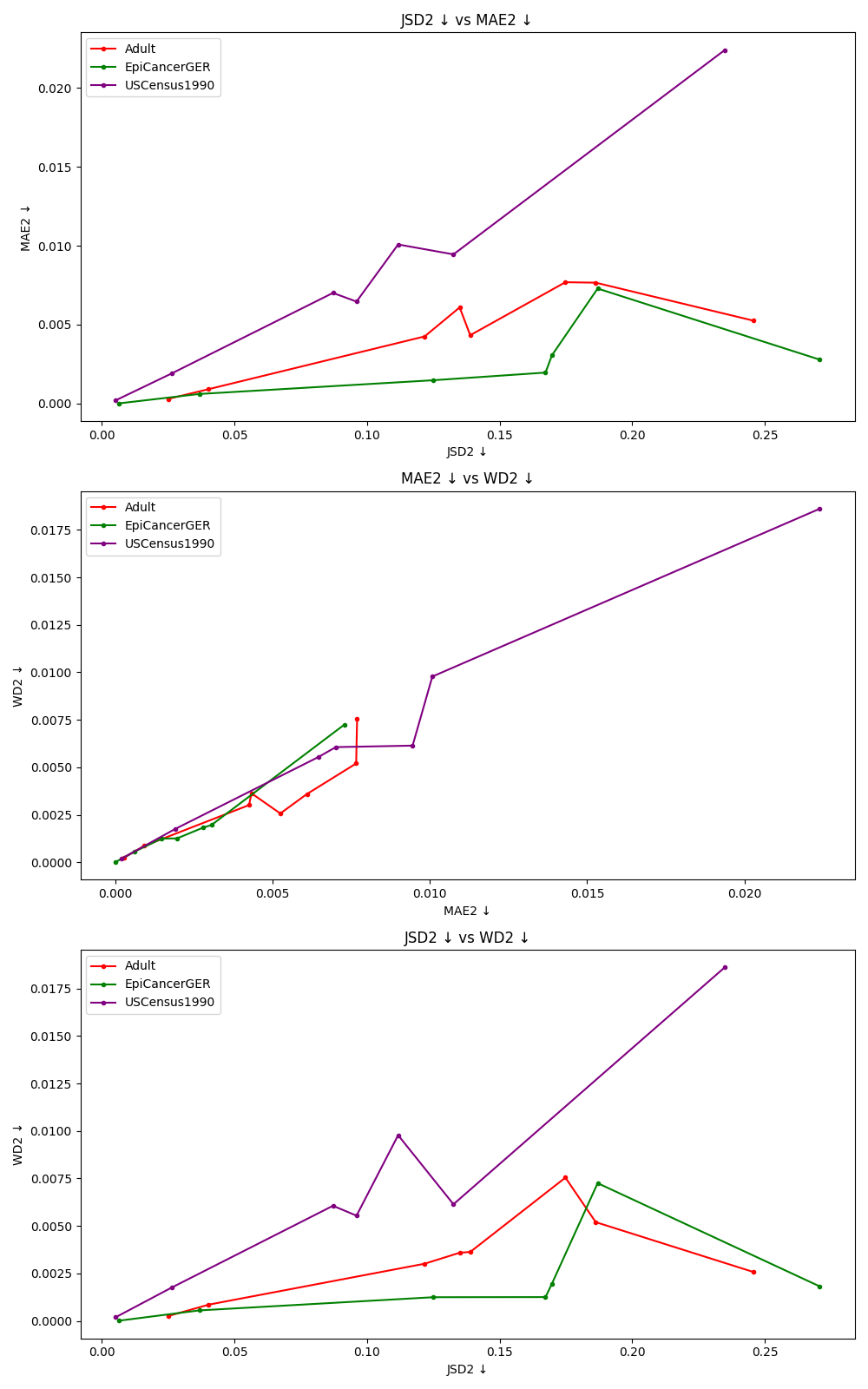}
  \caption*{Figure 5: Comparison of rankings from different metrics.}
  \label{fig:Figure 5}
\end{figure*}

\clearpage
\subsection{Variables included in EpiCancerGER dataset}
\label{sec:org9fd39ee}

\begin{table*}[h!]
  \centering
    \begin{tabular}{|l|p{0.222777\textwidth}|p{0.5777\textwidth}|}
\toprule
Variable & Label & Description \\
\midrule
SEX & Gender & Gender details:
1 = male
2 = female \\[0.777cm]
ALTGRP & Age group at diagnosis & The age at diagnosis categorized in 5-year age groups based on the (imputed) age at diagnosis:

a00b04 - from 0 to 4 years
a05b09 - from 5 to 9 years
…
a85plus - 85 years and above \\[1.777cm]
ICDGM10 & ICD-10-GM (GM=German Modification) & ICD code for the tumor diagnosis (4-digit, 10th Revision, WHO)
Valid codes:
C00.0 - C97.0 (excluding C77.0 - C79.9) (Malignant neoplasms without metastases)
D00.0 - D09.9 (In situ neoplasms)
D32.0 - D33.9 (Benign brain tumors)
D37.0 - D48.9 (Neoplasms of uncertain or unknown behavior)  \\[2.222777cm]
ICDGM10DREI & ICD-10-GM Chapter & ICD chapter for the tumor diagnosis (3-digit, 10th Revision, WHO) \\[0.777cm]
UICC & TNM to UICC  & Conversion of TNM classifications to UICC stages (for TNM editions 6 and 7) \\[0.777cm]
DIG & Dignity & Indication to the dignity:
0 = Benign neoplasm
1 = Neoplasm of uncertain or unknown behaviour
2 = In situ neoplasm
3 = Malignant neoplasm \\[1.777cm]
GRAD & Grading & Histopathological grading for solid tumors (1-digit):
1 = well differentiated
2 = moderately differentiated
3 = poorly differentiated
4 = undifferentiated/anaplastic
5 = low grade (low malignant)
6 = intermediate grade
7 = high grade (highly malignant) \\[1.777cm]
DSICH & Type of diagnostic confirmation & Indication of type of diagnostic confirmation (1 digit):
0 = autopsy
1 = clinical without spec. diagnostics
2 = clinical diagnosis
3 = death certificate (DCO)
4 = spec. tumor maker
5 = cytology
6 = Histology metastasis
7 = Histology primary tumor
8 = Other \\[1.777cm]
GROBST & Stage (internal use) & 1=in situ
2=local
3=regional
4=remote metastases
5=systemic disease
-999=unknown \\[0.777cm]
ZTYP & Cell Type & Cell type of  haemato-oncological malignancies (1-digit):
5 = T cell type
6 = B cell type
7 = Zero cell type
8 = Natural killer cell type \\[1.777cm]
\bottomrule
\end{tabular}

  \caption*{Table 8: Variables included in EpiCancerGER dataset.}
  \label{tab:Table 8}
\end{table*}

\clearpage
\subsection{Sex vs ICD 3-digit domain violations (EpiCancerGER dataset)}
\label{sec:org4b6076a}

\begin{table*}[h!]
  \centering
  \resizebox{\textwidth}{!}{
    \begin{tabular}{|l|l|l|}
\toprule
Female & Male & ICD Chapter \\
\midrule
C51 &   & Malignant neoplasm of vulva \\
C52 &   & Malignant neoplasm of vagina \\
C53 &   & Malignant neoplasm of cervix uteri \\
C54 &   & Malignant neoplasm of corpus uteri \\
C55 &   & Malignant neoplasm of uterus, part unspecified \\
C56 &   & Malignant neoplasm of ovary \\
C57 &   & Malignant neoplasm of other and unspecified female genital organs \\
C58 &   & Malignant neoplasm of placenta \\
D06 &   & Carcinoma in situ of cervix uteri \\
D39 &   & Neoplasm of uncertain or unknown behaviour of female genital organs \\
  & C60 & Malignant neoplasm of penis \\
  & C61 & Malignant neoplasm of prostate \\
  & C62 & Malignant neoplasm of testis \\
  & C63 & Malignant neoplasm of other and unspecified male genital organs \\
  & D40 & Neoplasm of uncertain or unknown behaviour of male genital organs \\
\bottomrule
\end{tabular}

  }
  \caption*{Table 9: ICD-10 Chapters considered for the domain violations analysis Sex vs ICD 3-digit.}
  \label{tab:Table 9}
\end{table*}

\clearpage
\subsection{Hyperparameters ranges}
\label{sec:org5c424eb}

\begin{table*}[h!]
  \centering
  \resizebox{\textwidth}{!}{
    \begin{tabular}{|l|l|r|r|r|}
\toprule
{} & {} & {\bfseries Processed as} & \multicolumn{2}{c|}{\bfseries Values} \\
{} & {} & {\bfseries } & {\bfseries min} & {\bfseries max} \\
{Model} & {Hyperparameter} & {} & {} & {} \\
\midrule
\multirow[t]{9}{*}{CTGAN} & batch\_size & optimized & 8 & 2048 \\
 & discriminator\_decay & optimized & 1e-07 & 0.1  \\
 & discriminator\_dim & stratified & (256, 256) or (512, 512) &  \\
 & discriminator\_lr & optimized & 1e-06 & 0.1  \\
 & embedding\_dim & optimized & 32 & 4096  \\
 & epochs & optimized & 10 & 1000  \\
 & generator\_decay & optimized & 1e-07 & 0.1  \\
 & generator\_dim & stratified & (128, 128) or (256, 256)  &  \\
 & generator\_lr & optimized & 1e-06 & 0.1  \\
\midrule
\multirow[t]{7}{*}{TVAE} & batch\_size & optmized & 32 & 4096  \\
 & compress\_dims & stratified & (128, 128) or (256, 256)  &  \\
 & decompress\_dims & stratified & (128, 128) or (256, 256) &  \\
 & embedding\_dim & optimized & 32 & 1024  \\
 & epochs & optimized & 10 & 200  \\
 & l2scale & optimized & 1e-07 & 0.1  \\
 & loss\_factor & optimized & 1.0 & 10.0  \\
\midrule
\multirow[t]{8}{*}{PGM} & 2\_cliques & fixed & 30 &  \\
 & 3\_cliques & fixed & 30 &  \\
 & bi\_nums & fixed & 30 &  \\
 & delta & fixed & 0.5000 &  \\
 & epsilon & fixed & 10000000.0 &  \\
 & max\_bins & optimized & 10 & 1000  \\
 & num\_iters & optimized & 1000 & 10000  \\
 & tri\_nums & fixed & 30 &  \\
\midrule
\multirow[t]{4}{*}{PrivSyn} & delta & fixed & 0.5 &  \\
 & epsilon & fixed & 100000000.0 & \\
 & max\_bins & optimized & 10 & 1000  \\
 & update\_iterations & optimized & 10 & 1000  \\
\midrule
\multirow[t]{7}{*}{TabDDPM} & batch\_size & optimized & 8 & 4096  \\
 & d\_layers & stratified & (256 x 7, 128) or (1024 x 7, 128) &  \\
 & dropout & optimized & 1e-06 & 0.7  \\
 & lr & optimized & 1e-05 & 0.1  \\
 & num\_timesteps & optimized & 10 & 10000 \\
 & steps & optimized & 10 & 100000  \\
 & weight\_decay & optimized & 1e-06 & 0.2  \\
\midrule
\multirow[t]{7}{*}{TableDiffusion} & batch\_size & optimized & 32 & 400  \\
 & d\_layers & stratified & (128, 128) or (256, 256)  &  \\
 & diffusion\_steps & optimized & 1 & 16  \\
 & epoch\_target & optimized & 1 & 16  \\
 & epsilon\_target & fixed & 100000000.0 &  \\
 & lr & optimized & 1e-05 & 0.1  \\
 & predict\_noise & optimized & False & True   \\
\bottomrule
\end{tabular}

  }
  \caption*{Table 10: Details of hyperparameters ranges. 'optimized' denotes hyperparameters tuned during HPO. 'stratified' means different HPO were run (only the bests are presented in the results).}
  \label{tab:Table 10}
\end{table*}

\end{document}